# Evolutionary Recurrent Decision Model in Developing Adaptive and Maladaptive Behaviors

Andrew Hu
andrewhu1997@gmail.com

*Abstract*—This study introduces the evolutionarily recurrent decision model (ERDM), a computational reinforcement learning framework designed to examine how evolutionary mismatch, bounded rationality, and satisficing contribute to adaptive and maladaptive behavior. ERDM simulates agents across evolutionary recurrent environments, including threat, prey/goal-pursuits, and alliances. Agents learn through competing rewards abstracted from survival metrics. A validity study under varying adverse childhood experiences demonstrates that distinct adaptive and maladaptive strategies, such as learned helplessness, avoidance, healthy relationships, and aggression, emerge naturally without being hardwired. These results align with empirical literature, showcasing ecological validity. The results suggest that many psychopathology-relevant aspects may be interpreted as bounded cognitive systems operating under modern-ancestral environmental mismatch, positioning ERDM as a key computational cognitive tool that can be extended to other studies.

## I. Introduction

Understanding the causes and reasons of mental health disorders is central to the field of psychology, psychiatry, and cognitive science. Evolutionary mismatch theory proposes a framework for understanding certain mental health disorders, stating that our psychological systems evolved in ancestral environments that are different than modern environments, leading to poor outcomes in our mental health, such as depression, anxiety, and stress-related disorders [1], [2], [3]. For example, modern-day chronic stressors, including work-related or media exposure, differ in quality and intensity from ancestral stressors of the environment and predators, and this difference leads to an increase in chronic disease and a rise in mental health disorders nowadays [1].

Evolutionary mismatch theory is further evidenced by the fact that our anatomically modern Homo Sapiens have evolved biologically and genetically through natural selection in conditions around 200,000 years ago, making it unlikely that our genotype has changed significantly. On the other hand, the modern lifestyle is remarkably different when compared to our ancestral hunter-gatherer lifestyle, outpacing the evolution of our cognitive mechanisms [4]. These environmental changes were contributed to by cultural, technological, and informational evolution, generating environments that differ significantly from the ancestral conditions that our cognition evolved from. This mismatch has been increasingly implicated in the rise of maladaptive behaviors and mental health disorders; however, understanding these underlying cognitive processes remains difficult to observe [1], [2], [4], [5]. This remains a critical challenge that computational cognitive approaches are positioned to address [5].

This study examines how cognition is constrained by bounded rationality and satisficing in the biological brain trained on ancestral environments, which can provide insight into the emergence of maladaptive behaviors and mental health disorders in modern living conditions [3], [6]. By modeling cognition as an adaptive, environment-dependent process within modern environmental constraints, this work also examines how these cognitive processes are shaped by their interaction and embedding in the environment. This study will use a new cognitive computational model, the Evolutionarily Recurrent Decision Model (ERDM), to conduct simulations under matched and mismatched ancestral conditions, thereby avoiding the ethical and practical limitations of real-world experimentation [7]. ERDM implements an adaptive single-agent-based modeling with deep learning capabilities within a stochastic, simulation-based framework to investigate complex cognitive and behavioral dynamics difficult to capture in closed-form or analytical models [5], [8], [9].

Rather than utilizing reinforcement learning to solve for an optimal policy for a predefined task, ERDM is designed to investigate how bounded rationality operating under competing evolutionary recurrent (survival) pressures, which are recurring environmental challenges ancestors faced, gives rise to context-dependent behavioral strategies [10]. Because these survival objectives are inherently in conflict, there is no universally optimal behavioral policy across environments, allowing adaptive and maladaptive behaviors to emerge naturally as environmental conditions change. Furthermore, ERDM is trained on simulated evolutionarily derived adaptive problems (problems that commonly occurred in ancestral times) with objective functions centered on survival pressures abstracted from physical fitness and social strategies of Dominance and Prestige [3], [11], [12]. The environment can be adjusted to emulate mismatched modern conditions, investigating the emergence of maladaptive and adaptive behaviors through decisions that satisfice these competing survival dimensions. The emergent behavioral outcomes of ERDM will be evaluated to determine how cognition is affected by the satisficing and bounded rationality from mismatched environments.

To evaluate whether ERDM captures behavior relevant to psychopathology, we examine agents exposed to varying levels of simulated adverse childhood experiences (ACEs), a domain supported by extensive literature that provides an external benchmark for validation and contextual information to interpret the model's outcomes. Empirical literature demonstrates that ACEs influence reward processing, perceived control, social behavior, and later mental health outcomes [13], [14], [15]. We will examine if ERDM reproduces these well-established behavioral patterns to evaluate the ecological validity of the proposed framework. To study the impact of mismatched environments on maladaptive behavioral patterns, three conditions are simulated: a stable, low-adversity environment; a moderate-adversity environment; and a high-adversity environment characterized by significant adverse childhood experiences (ACEs). Behavioral outcomes will be compared to current empirical findings on ACEs exposure and associated mental health outcomes to examine the model's ecological and theoretical validity. By simulating environments that are different than ancestral conditions, ERDM will naturally impose bounded rationality and satisficing constraints on the agent, allowing evaluation of whether bounded rationality under evolutionary mismatch is sufficient to produce adaptive and maladaptive behavioral patterns observed in empirical studies.

## II. METHODS

ERDM was implemented in Python using PyTorch and Numpy. Figure 1 illustrates the structure of ERDM. Iteratively, the agent interacts with evolutionarily recurrent environments, selects an action, receives fitness-relevant rewards, and updates internal policies with reinforcement learning. These behaviors are analyzed along with contextual information to categorize emergent complex behavioral strategies. The following sections will detail this process further.

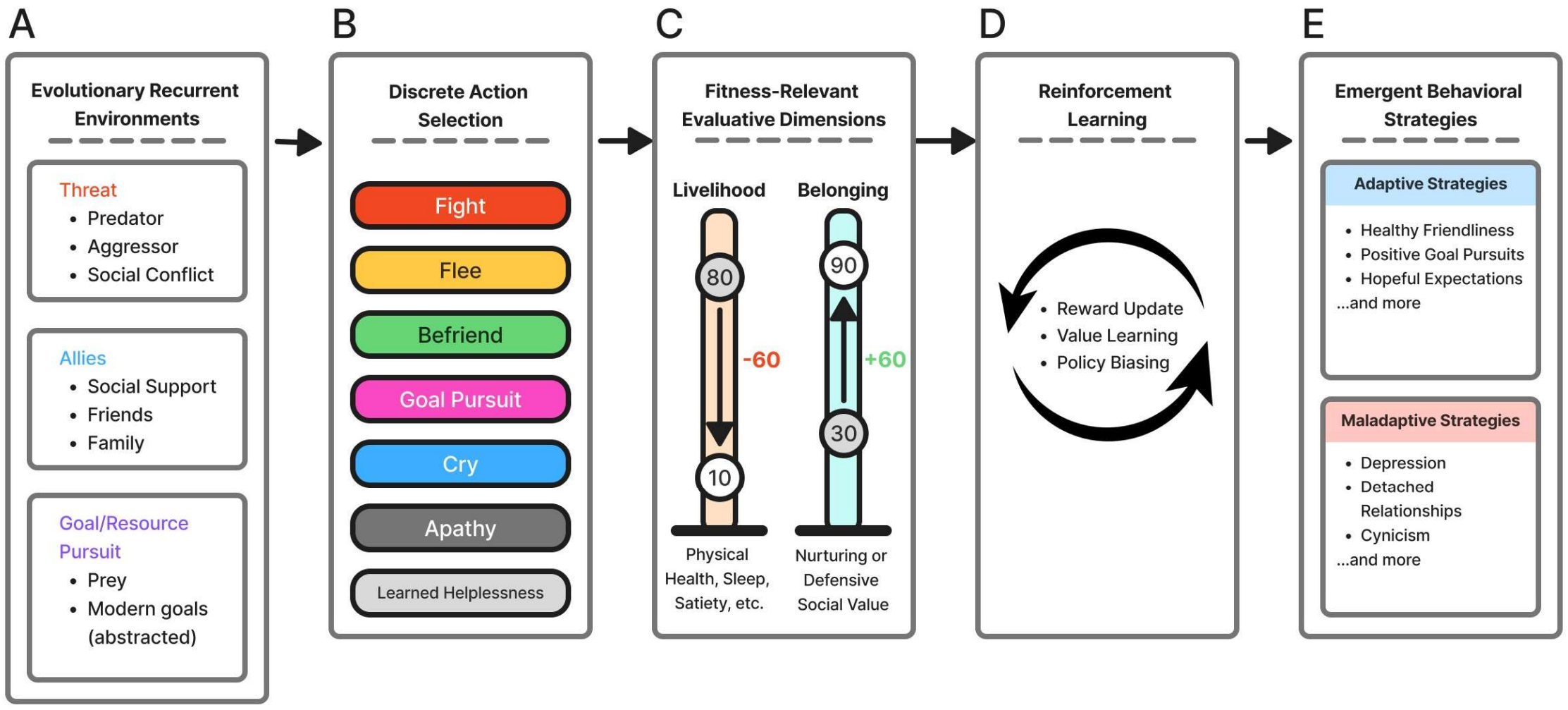

***Figure 1. ERDM utilizes discrete conserved environments and decisions.*** *(A) Evolutionary Recurrent Environment with examples (B) Agent chooses conserved decisions (C) Agent receives fitness-relevant evaluative rewards (D) Agent learns and updates policies, proceeds to next environment (E) Patterns of behaviors and reward expectations are organized as strategies*

### *A. Agent and Ecological Architecture*

The agent is a reinforcement learning system composed of multiple independent value neural networks, each corresponding to a behavioral action pathway. The agent will operate within a stochastic environment that represents discrete ecological events representing evolutionarily recurrent adaptive problems: threat, prey, and alliance opportunities (Figure 1).

Each network receives the internal and environmental state as inputs, and produces a scalar output representing the predicted reward for associating with the action in the current situation. Each action is associated with a separate value neural network $f_i(s; \theta_i)$, parameterized by weights $\theta_i$, which outputs a predicted reward value:

$$Vi_i(s) = f_i(s;\theta_i)$$

Where Vi(s) represents the expected reward if action $a_i$ is selected in state s. Action selection is conducted by choosing the action whose action network predicts the highest reward:

$$a* = \arg_i \max V_i(s)$$

To encourage exploration, action selection was implemented with an epsilon-greedy policy where, with probability $\epsilon$ a random action is selected and with probability $1 - \epsilon$ the greedy action a* is selected.

*B. Neural Network Structure*

Each action network consisted of a simple feedforward multilayer perception with two hidden layers. The architecture is consistent among all emotion/action pathways with the following structure:

- Input Layer: Size of State Representation
- Hidden Layer 1: 128 units with ReLU activation
- Hidden Layer 2: 64 units with ReLU activation
- Output Layer: 1 unit representing predicted reward.

Each simple feedforward network for emotional and behavioral action is to emulate how rudimentary cognitive mechanisms calibrated to ancestral pressure can produce complex behaviors.

*C. Action Selection and Reward Functions*

Complex behaviors are not explicitly encoded into the model and instead naturally emerge from patterns of repeated simple actions within context. The agent operates under three reward functions, one for each of the survival pressures of livelihood/physical fitness (L), defensive belonging/social value (DB), and nurturing belonging/social value (NB), which are calculated for each of the value networks for that situation-action pair (s, a). These two social values were abstracted from evolved dual strategies: dominance and prestige [11].

Reward functions are derived from causal consequences of the situation-action results. The rewards are calculated based on the environment's context, the agent's state, and the action. L rewards reflect physical consequences on an agent's well-being. DB rewards reflect the cost-inducing logic of dominance strategy, inflicting cost, or neutralizing threats across survival dimensions. In contrast, NB rewards reflect the benefit-providing logic of prestige strategy, generating positive values across the same dimensions [11], [16].

To illustrate the reward function framework, an example of a (s, a) threat-fight pair.

$$R_L(s,\text{fight}) = \min(L_{agent} - 1.1 \cdot L_{env}, 0)$$

$$R_{DB}(s,\text{fight}) = min\left(\frac{DB_{agent} + DB_{env}}{2} + \frac{L_{agent} - L_{env}}{2}\right) - DB_{agent}$$

$$R_{NB}(s,\text{fight}) = \min\left(-\frac{L_{env} - 1.1 \cdot L_{agent}}{L_{env}} \cdot NB_{env} \cdot 1.1, \ -5\right)$$

These equations showcase that a single action produces competing rewards through the three survival dimensions. For example, engaging in a fight with a threat may increase defensive belonging (e.g., dominance or protective power) but incurs losses in livelihood (physical harm) and nurturing belonging (capabilities of supporting others).

Due to the stochasticity of the environmental parameters, the state-dependent reward calculations, and multiple competing reward dimensions, the agent does not consistently converge to global optimal behavior nor memorize fixed action policies since no single action is universally optimized across all states or in a singular environment. Each action produces tradeoffs across multiple survival dimensions within that environment, forcing the agent to adopt satisficing strategies (locally sufficient rather than globally optimal) with the available information. These multi-objective and context-sensitive reward functions give rise not only to adaptive behaviors but also to maladaptive behaviors, constrained by bounded rationality and satisficing. Learned helplessness and apathy are action options, but the selection is based on internal reward expectation by the ERDM value network. These two state-dependent actions serve as the agent's refusal to commit to an action, either because all action selections are predicted to yield too negative a value (learned helplessness) or because none of the actions are predicted to yield rewards high enough to be considered rewarding (apathy).

*D. Episode Termination, Primary Belonging Value, and Reward Scaling*

Episode termination occurs when survival variables of livelihood/physical fitness (L) or a primary belonging dimension (DB or NB, depending on agent configuration) reach 0. This represents the agent's death or a failure of the social survival condition (e.g., social exile). An additional penalty is applied to the corresponding reward dimension that falls below the viability threshold:

$$R_i \leftarrow R_i - \lambda_{\text{death}}$$

where $R_i \in \{R_L, R_{DB}, R_{NB}\}$and $\lambda_{\text{death}}$is a fixed penalty of -20. Upon termination, the agent's state will reset to the initial state, and the simulation will continue. Furthermore, only the primary belonging dimension (DB or NB) that is designated for the agent's societal role will contribute to learning, while the non-primary belonging dimension is involved with the environment but does not influence the agent's reward optimization, creating different agent types (DB-dominant vs. NB-dominant). This selective prioritization showcases bounded rationality where agents optimize only a subset of survival signals and strategies.

Lastly, the reward functions are scaled using a piecewise function based on the agent's current state. This represents the greater importance of reward dimensions when diminished, capturing increased vulnerability under low-resource states. Each state variable is categorized into these discrete stages:

- **Dire** ($x < 10$)
- **Bad** ($10 \leq x < 20$)
- **Okay** ($20 \leq x < 50$)
- **Better** ($50 \leq x < 70$)
- **Stable** ($x \geq 70$)

A corresponding scaling factor is applied:

$$w(x) \in \{1.8, 1.5, 1.2, 1.1, 1.0\}$$

The final reward is computed as:

$$R_i = w(x_i^{\text{new}}) \cdot \Delta x_i$$

where:

- $\Delta x_i$ is the change in the survival variable
- $w(x_i^{\text{new}})$ increases sensitivity when the agent is in lower-resource states

*E. Simulating Adverse Childhood Events*

ERDM will be evaluated by simulating levels of adverse childhood experiences (ACE) to showcase whether ERDM reproduces plausible patterns of adaptive and maladaptive behaviors. A naïve model with no prior learning will be utilized to approximate early developmental learning conditions. In the ACE condition, threat environments are high-intensity events with environmental parameters for L, DB, and NB ranging from 80 – 100 (on a 0-100 scale). These values are high relative to the initial agent's survival parameters and the maximum value of 100 in L, DB, and NB. Meanwhile, the no ACE environments will have threats that are low intensity (20 – 40) that resemble nuisances or mild challenges rather than adverse events, and moderate-intensity environments (40-60) will have some difficult challenges. The frequency of the three evolutionarily adaptive problems is held constant across conditions. The internal reward, community strength, and risk parameters of the model will use the same default values.

Agents will be run with 200 independent simulations, each with 1500 training steps. Simulations are conducted with DB-dominant, NB-dominant, and random agents across three conditions: no ACE, moderate adversity, and ACE conditions. The random agent selects an action uniformly without considering state-context; however, learned helplessness and apathy are not in the random agent's action space since these actions require value estimates. The complex behavioral strategies emerging from these simulations will be analyzed using quantitative metrics, including frequency, action selection, internal state, reward expectations, and environmental state variables.

Hypotheses for this validation study are listed here:

- H1: Agents in the ACE group will have more actions of learned helplessness and apathy, reflecting the depressive results of adverse events.
- H2: Coping mechanisms expected to be unique to the agent's social role: NB-dominant agents (expected to contribute positivity to others) predicted to exhibit passive coping behaviors (fleeing, crying), whereas DB-dominant agents (expected to protect others) were predicted to exhibit active, sometimes aggressive behaviors, including misdirected aggression.

## III. RESULTS

### A. Learning Stabilization and Behavioral Diversity

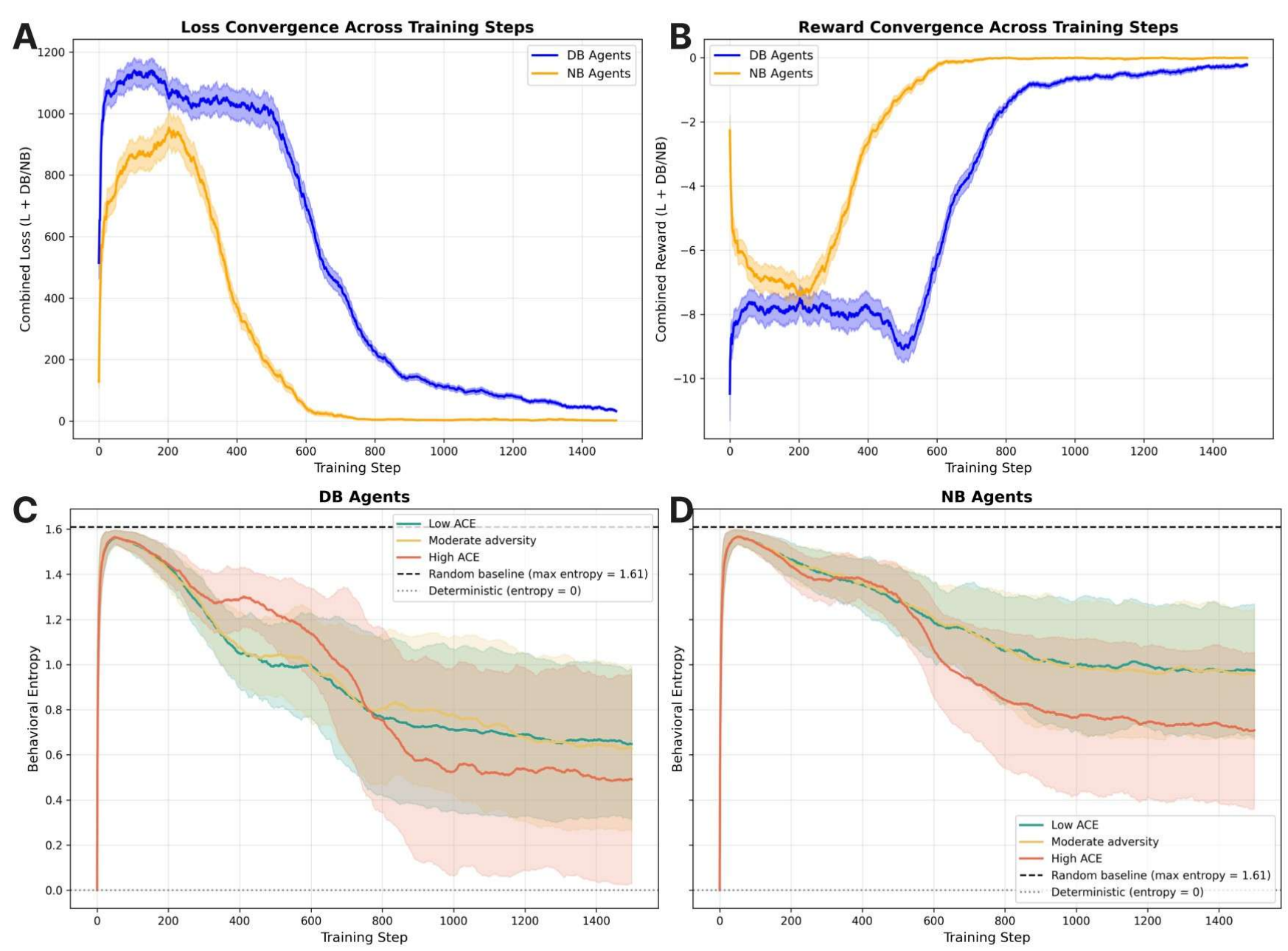

***Figure 2. Learning Stability and Behavioral Diversity.*** *(A) Loss convergence across training steps. (B) Reward stabilization across training steps. (C) Entropy of action selection across runs over training steps*

Value network loss decreased significantly from early to late training across all conditions and agent types. DB agents showed loss reduction ranging from 92% under No ACE to 97% under ACE. NB agents showed more pronounced learning stability as the loss reduced to over 99% across all conditions (Figure 2A). Late-window loss and reward slopes did not deviate from zero among all agents and conditions, indicating stable policy performance (Figure 2A, Figure 2B).

Behavioral entropy across runs was substantially smaller than loss reductions across all conditions. DB agents maintained between 34% and 44% of the initial entropy at late training, while NB agents maintained between 34% and 49% compared to the 92-99% loss reduction (Figure 2C, Figure 2D). Late-window slopes were not significantly different from zero across most conditions, indicating that behavioral diversity stabilized rather than declined. This dissociation between loss convergence and behavioral diversity was observed across all agent types and conditions.

The relationship between late-window loss and entropy was positive and significant across all conditions (Spearman's $\rho = 0.16$ to $0.88$, all $p < 0.05$). This relationship was stronger under ACE conditions for both DB ($\rho = 0.82$, $p < 0.001$) and NB agents ($\rho = 0.88$, $p < 0.001$) compared to No Ace and Moderate Adversity conditions. Under No Ace and Moderate Adversity conditions, behavioral entropy showed greater residual variability across runs despite the significant loss of convergence, showing weaker loss-entropy correlations (Figure 2C, Figure 2D). Under ACE conditions, the strong correlation indicates that loss convergence and behavioral consistency were more coupled.

*B. Complex Behavioral Strategy Emergence*

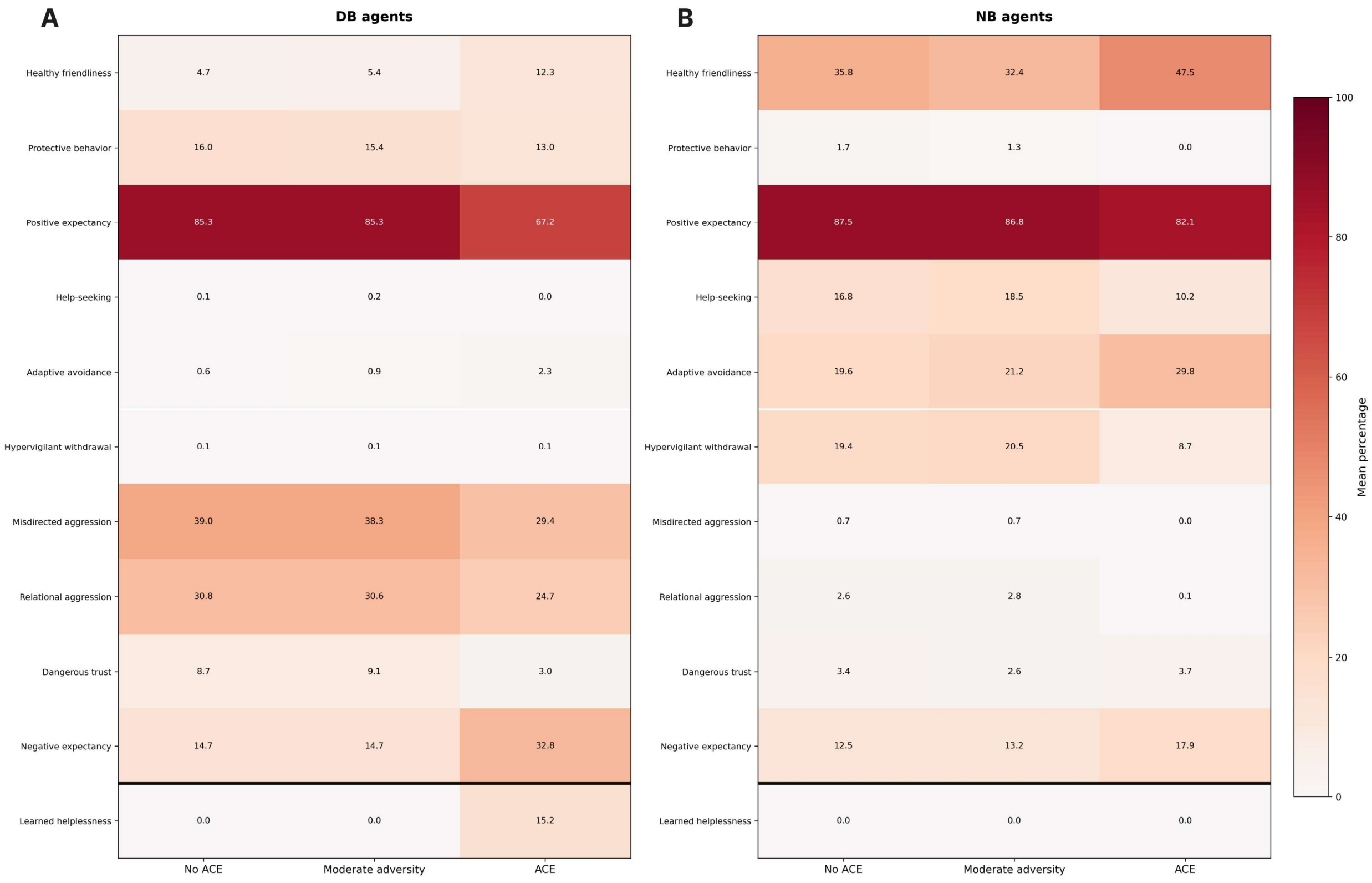

***Figure 3. Behavioral Policy Heatmap for DB agents (A) and NB agents (B)***

For Figure 3A, DB agents produced significant multivariate effects on adaptive behaviors (Wilks' $\lambda = .128$, $F(10, 1186) = 212.52$, $p < .001$) and maladaptive behaviors (Wilks' $\lambda = .081$, $F(12, 1184) = 247.88$, $p < .001$), with all four test statistics converging on this observation. According to the means, the profile was stable from No Ace to Moderate Adversity but changed significantly under ACE. In adaptive behaviors, positive expectancy notably declined from approximately 85% in No Ace and Moderate Adversity to 67% under ACE, and in maladaptive behaviors, helplessness was absent in both No Ace and Moderate Adversity to a mean of 15.2% under ACE. Negative expectancy in maladaptive behaviors also rose from ~15% to 33% under ACE. Notably, two of the externalizing behaviors, misdirected aggression (39% to 29.4%) and relational aggression (30.8% to 24.7%), declined modestly under ACE.

For NB agents (Figure 3B), conditions created significant multivariate effects on adaptive (Wilks' $\lambda = .321$, $F(10, 1186) = 90.81$, $p < .001$) and maladaptive behaviors (Wilks' $\lambda = .341$, $F(10, 1186) = 84.37$, $p < .001$). NB agents, unlike DB agents, had healthy friendliness as the primary behavior across all conditions and increased under ACE (~ 47%), while misdirected aggression and relational aggression remained near 0% and learned helplessness was absent entirely across conditions.

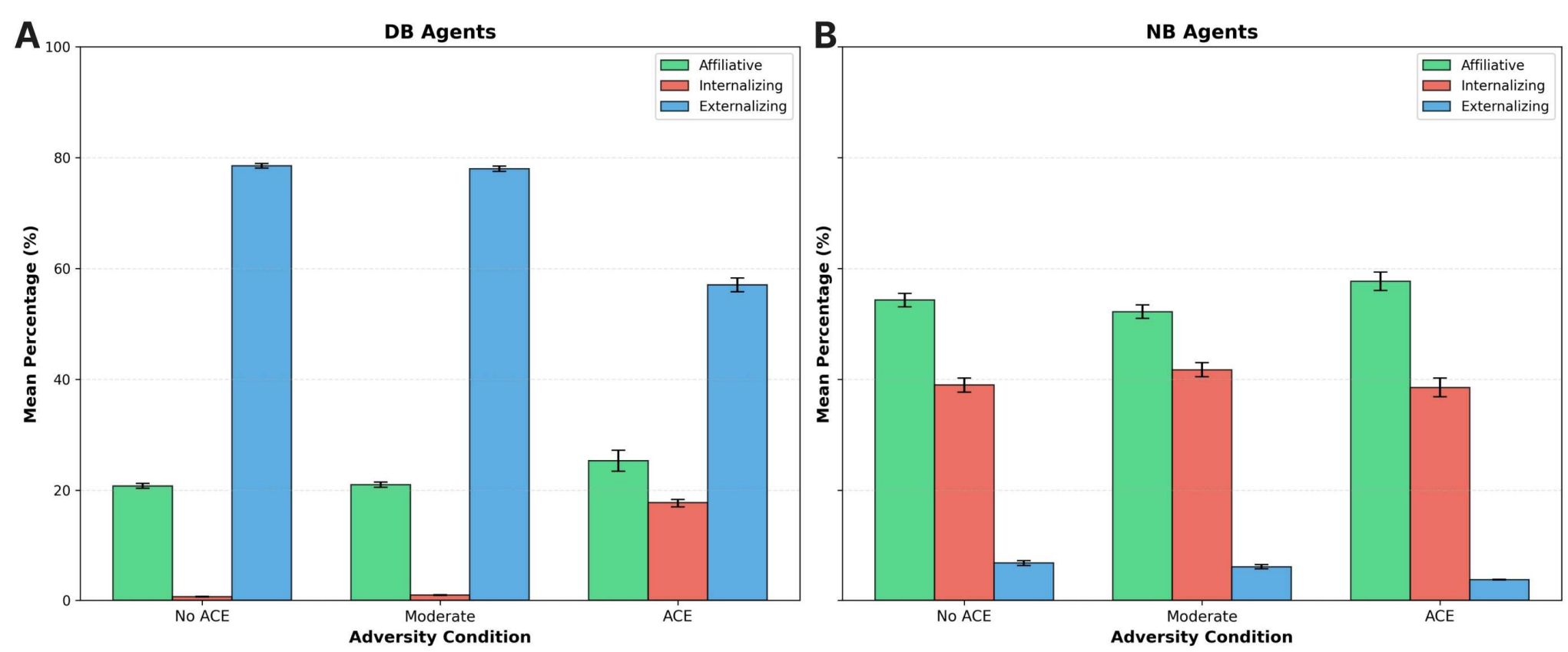

***Figure 4. Complex Behavioral Phenotype Bar Graphs of DB Agents (A) and NB Agents (B)***

In Figure 4A, Kruskal-Wallis tests revealed significant effects of conditions on DB agents' affiliative ($H$ = 30.05, $p < .001$), internalizing ($H = 369.34$, $p < .001$), and externalizing ($H = 308.65$, $p < .001$) clusters. Post hoc Dunn tests (Holm-corrected) indicated that these effects were exclusive to the ACE condition (DB agents under ACE differed significantly from No Ace and Moder Adversity across all three behavioral clusters, with $p < .001$). At the same time, the No ACE and Moderate Adversity did not differ with the affiliative or externalizing behaviors ($p > .60$), except for internalizing behaviors, which had a small but significant increase from No ACE to Moderate Adversity ($p < .001$). Under ACE, DB agents showed a sharp rise in internalizing behaviors (M = 17.6%, *SD* = 10.4%) relative to near-zero levels in non-ACE conditions (~0.7–1.0%), alongside a corresponding decline in externalizing behaviors (M = 57.1%, *SD* = 17.7% vs. ~78–79%).

In Figure 4B, NB agents had no variation in internalizing or affiliative clusters across conditions (both $p > .24$). Externalizing behaviors did show a significant overall effect ($H = 66.32$, $p < .001$) driven by ACE versus No ACE and ACE versus Moderate Adversity (both $p < .001$), with the ACE condition producing lower externalizing behaviors (M = 3.8%, *SD* = 0.90%) relative to the other two conditions (~6–7%).

Differences in the three complex behavioral phenotypes were significant between the agent type and all three conditions (all Mann-Whitney $p < .001$). NB agents showed higher affiliative (~52–58% vs. ~21–25%) and internalizing (~39–42% vs. ~0.7–17.6%) scores, while DB agents showed markedly higher externalizing scores (~57–79% vs. ~4–7%) across all conditions.

*C. DB-Dominant Agents Resilience with ACE*

Additional analysis was conducted on the DB-dominant agents under ACE conditions (with 1000 runs) to examine the heterogeneous complex behavior strategies that arise from the ACE condition. Figure 6A displays the distribution of the proportion of learned helplessness and the density through a histogram with KDE. The dataset strongly favored a two-component solution over a single-component model (ΔAIC ≈ 1173; ΔBIC ≈ 1158), indicating that the data are better described by two latent subpopulations rather than a unimodal distribution by a Gaussian mixture model. The two components were separated by a threshold (2.04% learned helplessness in a run) calculated by posterior probabilities from the mixture model to classify the latent subgroups of resilient (mean = 0.43%, n = 296) and vulnerable (mean = 20%, n = 704), with a clear distinction based on the separation metric of S = 2.50. The threshold is marked by the red dashed line in 6A, and Figure 6B shows the distribution across all runs within the aggregate, resilient, and vulnerable subgroups.

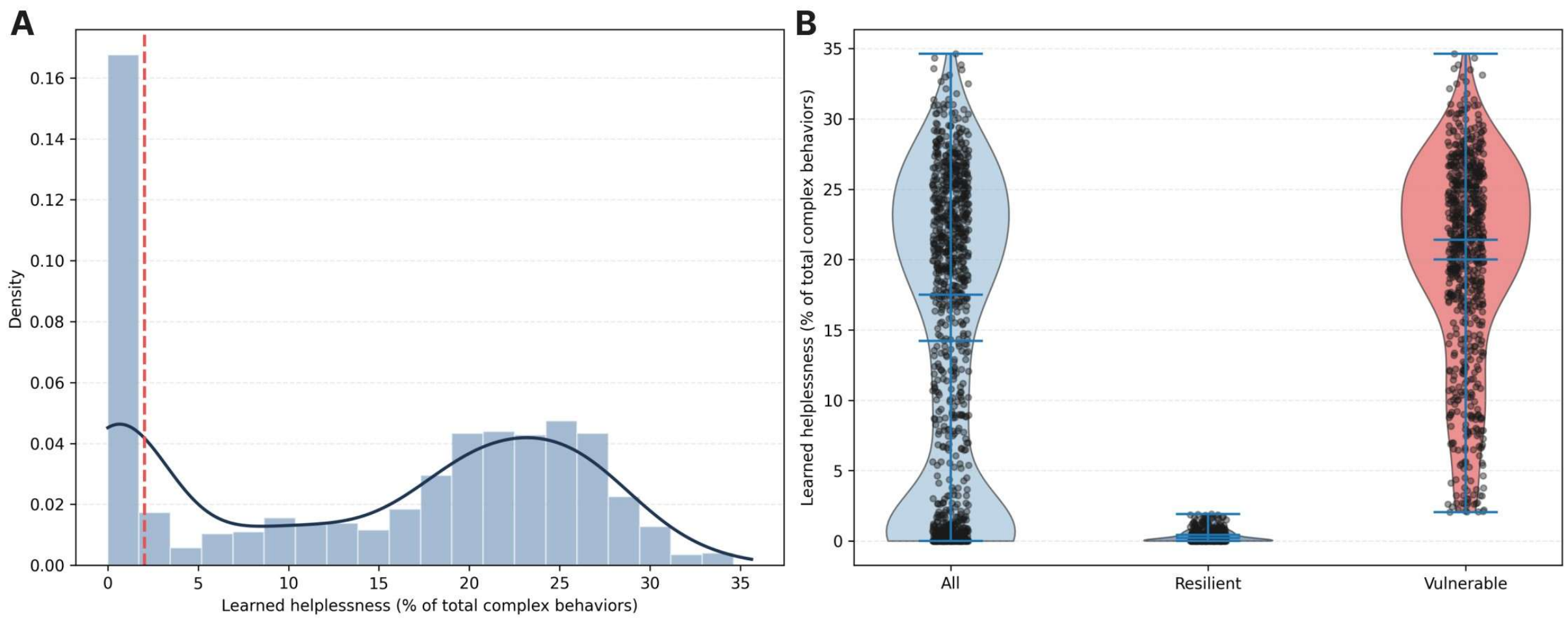

***Figure 5. Resilient vulnerable latent subgroups reveal themselves among learned helplessness distribution. (A) Histogram of Learned Helplessness with KDE. (B) Violin Plot of Latent Subgroup Split***

*D. Adaptive Behavior Emergence by Resilient Vulnerable Groups*

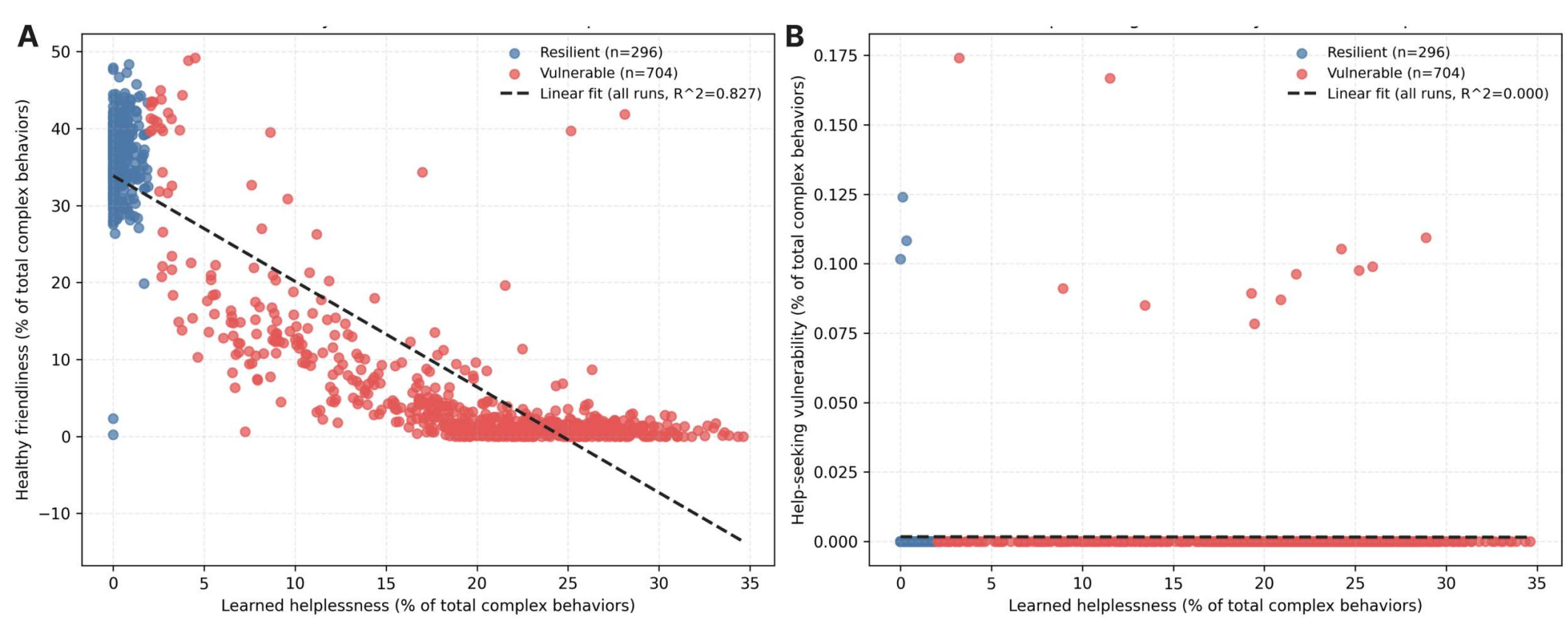

***Figure 6. Analysis of protective factors against learned helplessness. (A) Healthy Friendliness Plotted Against Learned Helplessness. (B) Help-Seeking Vulnerability Plotted Against Learned Helplessness***

Known protective factors (healthy relationships/friendliness and help-seeking vulnerability, Figure 6) against ACE and stressors were analyzed to investigate the mechanisms underlying resilience in these agents [17], [18]. In Figure 6A, linear regression was performed among all agents to find that healthy friendliness demonstrated a strong negative relationship with learned helplessness (Pearson $r = -0.91$, $R^2 = 0.83$, $p < 0.001$; Spearman $\rho = -0.84$, $p < 0.001$), signifying the strong influence healthy relationships have on resilience against ACE. Consistent with this, resilient agents displayed significantly higher levels of healthy friendliness when compared to vulnerable agents (Mann-Whitney U, $p < 0.001$).

In contrast, help-seeking vulnerability showed no significant association with learned helplessness (Pearson $r \approx 0$, $p = 0.91$; Spearman $\rho \approx 0$, $p = 0.87$) in Figure 6B, and no difference was observed between the resilient and vulnerable subgroups.

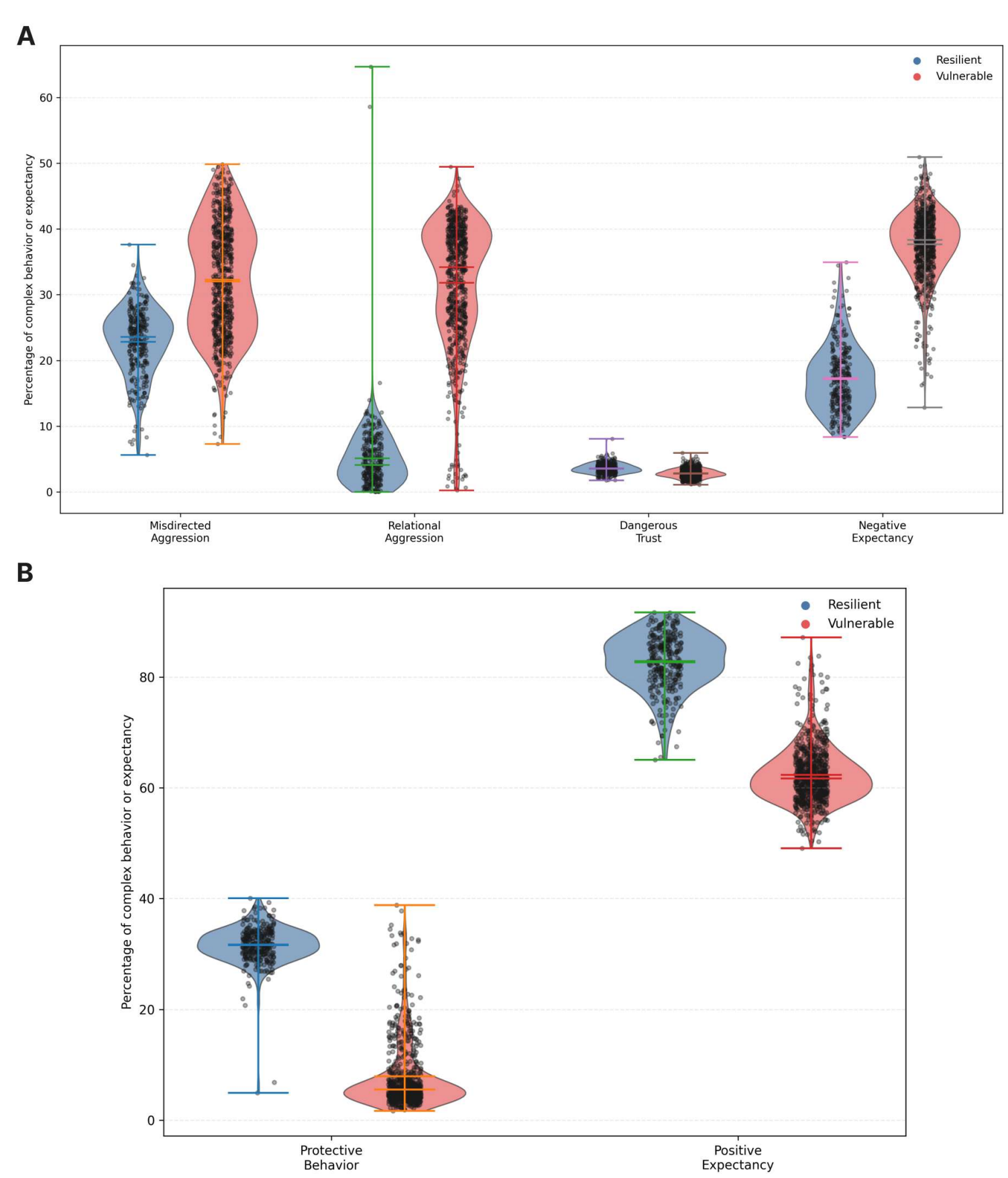

***Figure 7. Complex Behaviors Emerge Uniquely Between Resilient and Vulnerable Agents*** *(A) Adaptive behavior emergence comparison between resilient and vulnerable groups. (B) Maladaptive behavior emergence comparison between resilient and vulnerable groups.*

A multivariate analysis of variance (MANOVA) revealed a significant effect of these two latent subgroups on adaptive complex behavioral strategies (Wilks' $\lambda = 0.185$, $F(2, 997) = 2197.17$, $p < 0.001$), signifying a strong distinction between resilient and vulnerable agents through a multivariate complex behavior space (Figure 7A). Follow-up univariate analyses using Mann-Whitney U tests indicated that these differences existed across the behaviors. Resilient agents exhibited higher levels of protective behavior (mean = 31.64% vs. 7.92%; Cohen's $d = -4.23$, $p < 0.001$) and positive expectancy (mean = 82.65% vs. 62.35%; Cohen's $d = -3.94$, $p < 0.001$), with effects surviving multiple comparison correction (Figure 7A).

Similarly, maladaptive complex behaviors were also analyzed by MANOVA, revealing a robust multivariate effect of these latent subgroups on maladaptive complex behaviors (Wilks' $\lambda = 0.243$, $F(3, 996) = 1036.13$, $p < 1.2 \times 10^{-305}$), indicating a strong distinction between resilient and vulnerable agents in maladaptive complex behaviors (Figure 7B). Follow-up univariate analyses confirmed that maladaptive behaviors significantly differed between groups. Misdirected aggression (vulnerable: mean = 32.29%, resilient: mean = 22.83%, Mann–Whitney U = 38,607, $p = 9.35 \times 10^{-56}$, Cohen's d = 1.24), relational aggression (vulnerable: mean = 31.83%, resilient: mean = 5.12%, U = 5,874.5, $p < 10^{-123}$, Cohen's d = 3.16), and negative expectancy bias (vulnerable: mean = 37.65%, resilient: mean = 17.35%, U = 2,309, $p < 10^{-131}$, Cohen's d = 3.94) showed a pronounced increase in vulnerable agents when compared to resilient agent. In contrast, dangerous trust exhibited a smaller, opposite effect, with resilient agents having slightly higher levels than vulnerable agents (resilient: mean = 3.61%, vulnerable: mean = 2.85%, U = 160,898, $p = 3.96 \times 10^{-42}$, Cohen's d = -1.04).

## IV. Discussion

### A. *Model Validity and Emergent Behaviors*

A central finding in the model's validity is the dissociation between learning convergence and behavioral convergence. Reinforcement learning loss stabilized across all conditions and agent types, showcasing reliable learning from environmental feedback (Figure 2A, Figure 2B). On the contrary, behavioral policies did not converge on a single optimum; instead, agents adopted diverse, stable strategies across runs (Figure 2C, Figure 2D). This dissociation reflects real-world dynamics that even when exposed to identical environmental pressures, organisms do not converge to uniform behavioral strategies. For these agents, diverse strategies are shaped by stochastic experience and recent reward history.

The emergence of many behavioral strategies without explicit encoding supports internal validity of ERDM as a model of cognitive adaptation rather than scripted or guided behavioral outputs by these survival reward functions. Notably, behavioral entropy was more strongly associated with loss convergence under ACE conditions, suggesting that high adversity narrows available behavioral options. This pattern may reflect how high-threatening environments constrain the viability of diverse adaptive strategies (Figure 2).

### B. *Maladaptive Inactivity as Rational Survival Inhibition*

The first hypothesis was partially supported. Learned helplessness emerged exclusively under ACE conditions in DB-dominant agents, while apathy was absent across all conditions. This dissociation of learned helplessness and apathy occurrences was not predetermined by the reward architecture, as both learned helplessness and apathy represent forms of inaction that the model does not favor one more than the other.

Within the ERDM framework, this dissociation showcases two distinct mechanisms underlying inactivity. Learned helplessness is a risk-focused inactivity that results when all available actions are predicted to yield negative outcomes beyond the agent's risk threshold, a rational inhibition of action under overwhelming risk [19]. By contrast, apathy is a reward-focused inactivity that reflects a failure to find sufficient reward value in any available action space, representing a motivationally depleted state. That ACE conditions produced learned helplessness rather than apathy suggests that under chronic adversity, the agent will be focused predominantly on risk evaluation rather than reward motivation.

This reframes learned helplessness not as a pathological failure of reward processing but as a computationally logical result of a survival system encountering overwhelming risks that the agent cannot find suitable adaptive strategies against [19]. This interpretation aligns with and extends empirical findings associating childhood threats/abuse with threat hypervigilance or risk appraisal (learned helplessness) rather than anhedonia (apathy), which is more commonly found in neglect [20]. More empirical studies could investigate the utility of interventions that address threat or risk appraisal mechanisms rather than motivational systems depending on which state underlies the inactivity.

### *C. Ecological Validity of NB and DB-Dominant Agents*

The diverging behavioral profiles of NB and DB-dominant agents (originating from the dual strategies theory of social hierarchies and value) closely mirror established internalizing and externalizing behaviors established in developmental psychopathology (Figure 4A, Figure 4B). Instead of being pre-defined, these behavioral profiles emerged from a single difference in prioritization of nurturing versus defensive social value, suggesting that internalizing/externalizing distinctions may be partially traceable in how individuals weigh prestige (NB) versus dominance (DB) survival strategies in their reward processing.

NB-dominant agents exhibited avoidance-based coping (including increased fleeing and reduced help-seeking) that intensified as adversity increased. This phenomenon occurs when allies of the agent are not capable enough to handle the threats, and so NB strategies of befriending allies incur fitness cost. As a result, avoidance-based behaviors become the locally optimal satisficing response rather than a more prosocial behavior. This transition aligns with the empirical findings that unresponsive or incapable caregiving produces avoidant attachment styles characterized by reduced reliance on others, leading to fewer help-seeking behaviors when the caregivers or community cannot protect them from the cost of these threats. However, internalizing behaviors remain stable across adversity levels as NB-dominant agents decreased avoidant behaviors and adopted a more empirically studied tend-and-befriend stress response with increasing threats. Tend-and-befriend is a well-studied alternative to fight-or-flight that seeks social affiliation when stressed, especially among individuals who prioritize nurturing strategies [21], [22]. In contrast, DB-dominant agents showed externalizing trajectories, including fight-or-flight responses [23], further supporting the ecological validity of the agent-type distinction through social strategies.

DB-dominant agents showed a significant compositional shift in maladaptive behavior under ACE conditions, consistent with the idea that high adversity would increase internalizing outcomes. Learned helplessness emerged exclusively under ACE, and negative expectancy rose sharply from ~15% to 33% (Figure 3A). Unexpectedly, both misdirected and relational aggression declined rather than escalating, reflecting a compositional shift in maladaptive behavior rather than a uniform increase (Figure 3A). One potential explanation is that DB agents' persistent baseline aggression towards non-threatening targets under low-to-moderate adversity reflects normal dominance-display behavior, which serves an evolutionary function to help establish hierarchy, rather than a sign of aberrant behavior [24].

However, this baseline dominance and hierarchy establishment becomes computationally untenable as threat intensity exceeds the agent's risk threshold, causing learned helplessness to displace the aggressive strategies. The resilience analysis supports this assessment as vulnerable DB agents retained significantly higher aggression relative to agents, suggesting the aggregate decline is due to the emergence of learned helplessness rather than a genuine reduction in dominance-oriented behavior (Figure 7A, 7B). Supplemental analyses further suggest that when allies are strong, DB-dominant agents shift from aggression toward affiliation, indicating that DB value can also be accrued through association with high-status allies. Utilizing this affiliation as an alternate method to establish hierarchy in turn lowers the baseline dominance display (Figure S1). Overall, these showcase mechanistic insights in presenting maladaptive behaviors not as a cognitive failure but as a rational strategy in the presence of survival pressure of mismatched environments.

### *D. Resilience and Vulnerability as Emergent Survival Phenotypes*

The emergence of resilient and vulnerable latent subgroups within DB-dominant agents under ACE conditions shows that behavioral variability can arise under identical environmental conditions and intensity (Figure 5B). Although the environmental parameters were held constant across runs, the stochastic ordering of early events shaped reward trajectories that diverged into distinct behavioral phenotypes.

The resilient latent subgroup was strongly associated with healthy friendliness, with a near-perfect negative relationship between social affiliation and learned helplessness ($r = -0.91$) (Figure 6A). This aligns with positive social relationships as a primary protective factor in trauma-exposed populations [25], [26]. Interestingly, the null finding for help-seeking vulnerability as a factor is a consequence of the DB reward strategy. DB-dominant agents deprioritize help-seeking because acts of vulnerability are appraised as detrimental to DB, signaling weakness, reducing perceived dominance capacity, and overall predicted to incur survival costs within the agent's social standings. According to empirical data, help-seeking can be experienced as threatening to self-esteem [23] and so generates a testable clinical hypothesis: for children with DB-dominant behavioral profiles, interventions targeting the development of positive social relationships are likely to be more effective than those focused on help-seeking skill training. This hypothesis, arising from the model, demonstrates the ERDM's usability in producing novel questions and intervention targets.

## V. Limitations

Several limitations should be considered with the current findings and the ERDM framework. First, the model assumes an accurate and unbiased perception of environmental states by the agent, whereas real-world cognition is shaped by distortions, cognitive biases, and self-referential miscalibration. Future iterations of ERDM could incorporate perceptual noise and self-image distortion features, allowing agents to overestimate or underestimate their own capabilities and environmental properties.

Second, the current framework simplifies interactions into only three evolutionarily recurrent situations (threat, prey/goal-pursuit, and alliance opportunities) and a handful of conserved actions. While this abstraction is an intended simplification to enable tractable modeling of adaptive problems, it does not capture the complexity of these dynamics, including empathy, failed efforts, and unique community values.

Third, the present model is a single-agent reinforcement learning framework, which limits the ability to capture emergent group dynamics and interactions with other dynamic agents. Human behavior in real-world contexts is fundamentally multi-agent, and so extending ERDM into a multi-agent framework would allow for the study of population-level phenomena such as cultural transmission and social evolution [9].

Finally, although the framework is designed to be translatable into empirical settings while revealing often unobservable underlying cognitive mechanisms, the study does not include an accompanying experimental validation. Therefore, the findings remain computational, and future research is needed to evaluate ERDM-behavioral predictions and suggested interventions.

## VI. Conclusion

In summary, ERDM demonstrates that complex adaptive and maladaptive behaviors can emerge from simple reinforcement learning agents operating under evolutionarily grounded constraints. The findings suggest that behavioral strategies arise not from randomness, but from interactions between environmental pressures, bounded rationality, and internal competing policy learning that brings about satisficing. Critically, behaviors such as learned helplessness, aggression, and avoidance/withdrawal may not reflect a pure intrinsic dysfunction, but rather the predictable outcomes of rational agents operating within highly adverse and/or evolutionarily mismatched environments. Highlighting a theoretical contribution to be explored of whether maladaptive behavior is not a failure of cognition, but the consequence of rational output from evolutionary survival metrics to a mismatched context.

Importantly, this study highlights ERDM as a flexible and extensible computational framework for ethical study of behaviors under a wide range of real-world conditions that can be mapped to these evolutionarily recurrent adaptive problems. While the present study focuses on adverse childhood experiences as a representative case, the same framework can be generalized to other modern contexts by modifying environmental parameters and agent properties. Even within this ACE validity study, only a subset of possible behavioral interpretations and latent mechanisms were explored, suggesting substantial space for further discovery.

Future extensions of ERDM could incorporate a broader range of adaptive problems and conserved actions, more detailed environment outcomes, perceptual and self-referential distortions, and multi-agent interactions. These additions would enable more nuanced modeling of how individuals adapt across different landscapes and developmental stages. Furthermore, this could greatly extend the real-world modeling that ERDM could analyze.

Overall, ERDM provides a foundational step towards a more mechanistic understanding of behavioral adaptation under evolutionary constraints. The strength lies not only in reproducing empirically observed patterns but in offering a framework that uncovers unobserved cognitive mechanisms, in which new hypotheses about cognition and behaviors can be systematically and ethically explored.

## Supplementary

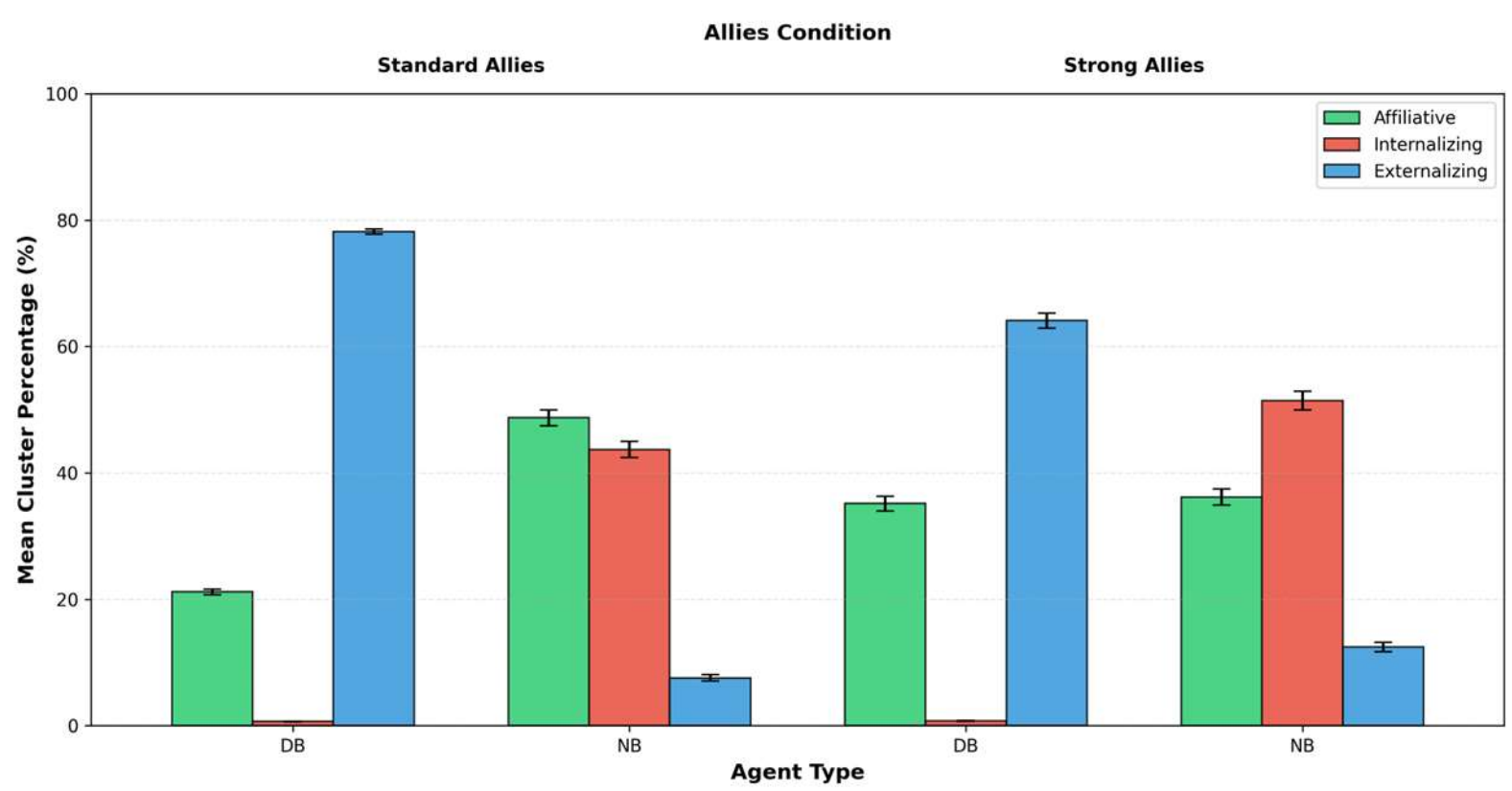

***Figure S1. Behavioral cluster composition under standard versus strong ally conditions at no ACE condition.*** *Mean percentage of total complex behaviors assigned to affiliative (healthy friendliness, community-trusting vulnerability, protective behavior), internalizing (hypervigilant withdrawal, adaptive avoidance, learned helplessness), and externalizing (misdirected aggression, relational aggression, dangerous trust) clusters for defensive-belonging (DB) and nurturing-belonging (NB) agents. Data are from 200 independent simulation runs per agent type and allies' condition. Bars show run-level means and error bars indicate SEM. Pairwise comparisons between standard and strong ally conditions within each agent type and behavioral cluster were conducted using two-sided Mann–Whitney U tests (n = 200 per group). For DB agents, strong ally conditions were associated with a significant increase in affiliative behavior and a significant decrease in externalizing behavior (both $p < 10^{-18}$), while internalizing behavior did not differ significantly ($p = 0.096$). In contrast, NB agents exhibited significant differences across all three behavioral clusters under strong ally conditions (all $p < 10^{-4}$ or smaller).*